%% file: neurips_2025.tex
\documentclass{article}

 \usepackage[dandb, final]{neurips_2025}
\usepackage[utf8]{inputenc} % allow utf-8 input
\usepackage[T1]{fontenc}    % use 8-bit T1 fonts
\usepackage{hyperref}       % hyperlinks
\usepackage{url}            % simple URL typesetting
\usepackage{booktabs}       % professional-quality tables
\usepackage{amsfonts}       % blackboard math symbols
\usepackage{nicefrac}       % compact symbols for 1/2, etc.
\usepackage{microtype}      % microtypography
\usepackage{xcolor}         % colors

\input{preamble}

\definecolor{cvprblue}{rgb}{0.21,0.49,0.74}

\usepackage{amsmath}
\usepackage{graphicx}
\usepackage{booktabs}
\usepackage{adjustbox}
\usepackage{multirow}

\title{MME: A Comprehensive Evaluation Benchmark \\ for Multimodal Large Language Models}

\author{%
\\
  Chaoyou Fu$^{1,2,\spadesuit}$, Peixian Chen$^{3}$, Yunhang Shen$^{3}$, Yulei Qin$^{3}$, Mengdan Zhang$^{3}$ \\ 
  Xu Lin$^{3}$, Jinrui Yang$^{3}$, Xiawu Zheng$^{4}$, Ke Li$^{3,\dagger}$, Xing Sun$^{3}$ \\ 
  Yunsheng Wu$^{3}$, Rongrong Ji$^{4}$, Caifeng Shan$^{1,2}$, Ran He$^{5}$ \\
  \\
  $^{1}$State Key Laboratory for Novel Software Technology, Nanjing University \\
  $^{2}$School of Intelligence Science and Technology, Nanjing University \\
  $^{3}$Tencent Youtu Lab \quad $^{4}$Xiamen University \quad $^{5}$CASIA 
  \and
    \footnotesize{
    $^{\spadesuit}$~Project Leader \;
    $^{\dagger}$~Corresponding Author \;}
}

\begin{document}
\maketitle
\input{sec/0_abstract}    
\input{sec/1_intro}

\input{sec/t_table}
\input{sec/2_data}

\input{sec/3_experiment}

\input{sec/4_analysis}

\input{sec/5_conclusion}

\section*{Acknowledgments}
This work is funded by National Natural Science Foundation of China (Grant No. 62506158 and No. 62441234), Fundamental Research Funds for the Central Universities, AI \& AI for Science Project of Nanjing University (No. 2024300529), and CCF-Tencent Rhino-Bird Open Research Fund.

{
    \small
    \bibliographystyle{plain}
    \bibliography{main}
}

\end{document}

%% file: preamble.tex
\usepackage[dvipsnames]{xcolor}

%% file: sec/0_abstract.tex
\begin{abstract}
Multimodal Large Language Model (MLLM) relies on the powerful LLM to perform multimodal tasks, showing amazing emergent abilities in recent studies, such as writing poems based on an image. However, it is difficult for these case studies to fully reflect the performance of MLLM, lacking a comprehensive evaluation. In this paper, we fill in this blank, presenting the first comprehensive \textbf{M}LL\textbf{M} \textbf{E}valuation benchmark \textbf{MME}. It measures both perception and cognition abilities on a total of 14 subtasks. In order to avoid data leakage that may arise from direct use of public datasets for evaluation, the annotations of instruction-answer pairs are all manually designed. The concise instruction design allows us to fairly compare MLLMs, instead of struggling in prompt engineering. Besides, with such an instruction, we can also easily carry out quantitative statistics. A total of 30 advanced MLLMs are comprehensively evaluated on our MME, which not only suggests that existing MLLMs still have a large room for improvement, but also reveals the potential directions for the subsequent model optimization. The data are released at the project page: \url{https://github.com/BradyFU/Awesome-Multimodal-Large-Language-Models/tree/Evaluation}.
\end{abstract}

%% file: sec/1_intro.tex
\section{Introduction}
\label{sec:intro}

The thriving of Large Language Model (LLM) has paved a new road to the multimodal field, i.e., Multimodal Large Language Model (MLLM) \cite{openai2023gpt4,huang2023language,li2023blip,driess2023palm,team2025kimi,tu2024overview}. 
It refers to using LLM as a brain to process multimodal information and give reasoning results \cite{yin2023survey}. 
Equipped with the powerful LLM, MLLM is expected to address more complex multi-modal tasks \cite{driess2023palm,wen2025dycrowd,shen2023hugginggpt,lin2023sphinx,zhang2025speechact,zhao2023mmicl,liu2023mmbench,liu2021geometry,zhang2025logavatar,ying2024mmt,fang2024mmbench}.The three representative abilities of LLM \cite{zhao2023survey}, including instruction following \cite{touvron2023llama}, In-Context Learning (ICL) \cite{brown2020language}, and Chain-of-Thought (CoT) \cite{wei2022chain} are also manifested in multimodality. 
For example, Flamingo \cite{alayrac2022flamingo} turns on multimodal ICL, which can adapt to new tasks by giving a few examples. 
PaLM-E \cite{driess2023palm} achieves amazing OCR-free math reasoning via CoT.
GPT-4V~\cite{openai2023gpt4} shows even more ability in a variety of complex reasoning tasks \cite{wu2023early}.
MiniGPT-4 \cite{zhu2023minigpt} implements GPT-4\cite{openai2023gpt4}-like instruction following capabilities, such as converting images into corresponding website codes, by introducing multimodal instruction tuning. These emergent abilities of MLLMs are exciting and imply that a new dawn has broken in artificial intelligence.

\begin{figure*}[t]
    \centering
    \includegraphics[width=1.0 \textwidth]{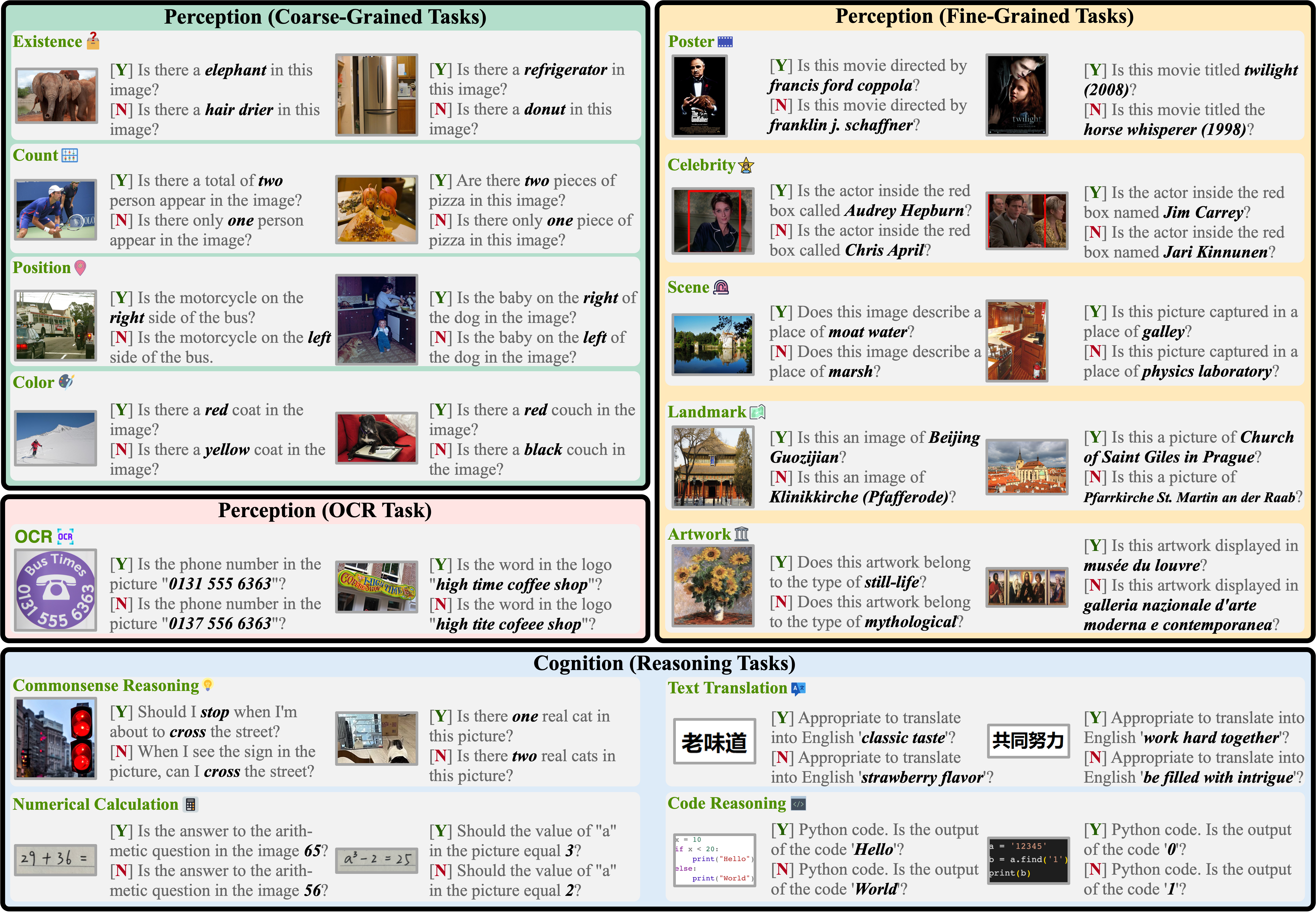}
    \caption{Diagram of our MME benchmark. It evaluates MLLMs from both perception and cognition, including a total of 14 subtasks. Each image corresponds to two questions whose answers are marked yes [Y] and no [N], respectively. The instruction consists of a question followed by ``Please answer yes or no''. It is worth noting that all instructions are manually designed.}
    \label{fig-dataset}
\end{figure*}

\begin{figure*}[t]
    \centering
    \includegraphics[width=0.95 \textwidth]{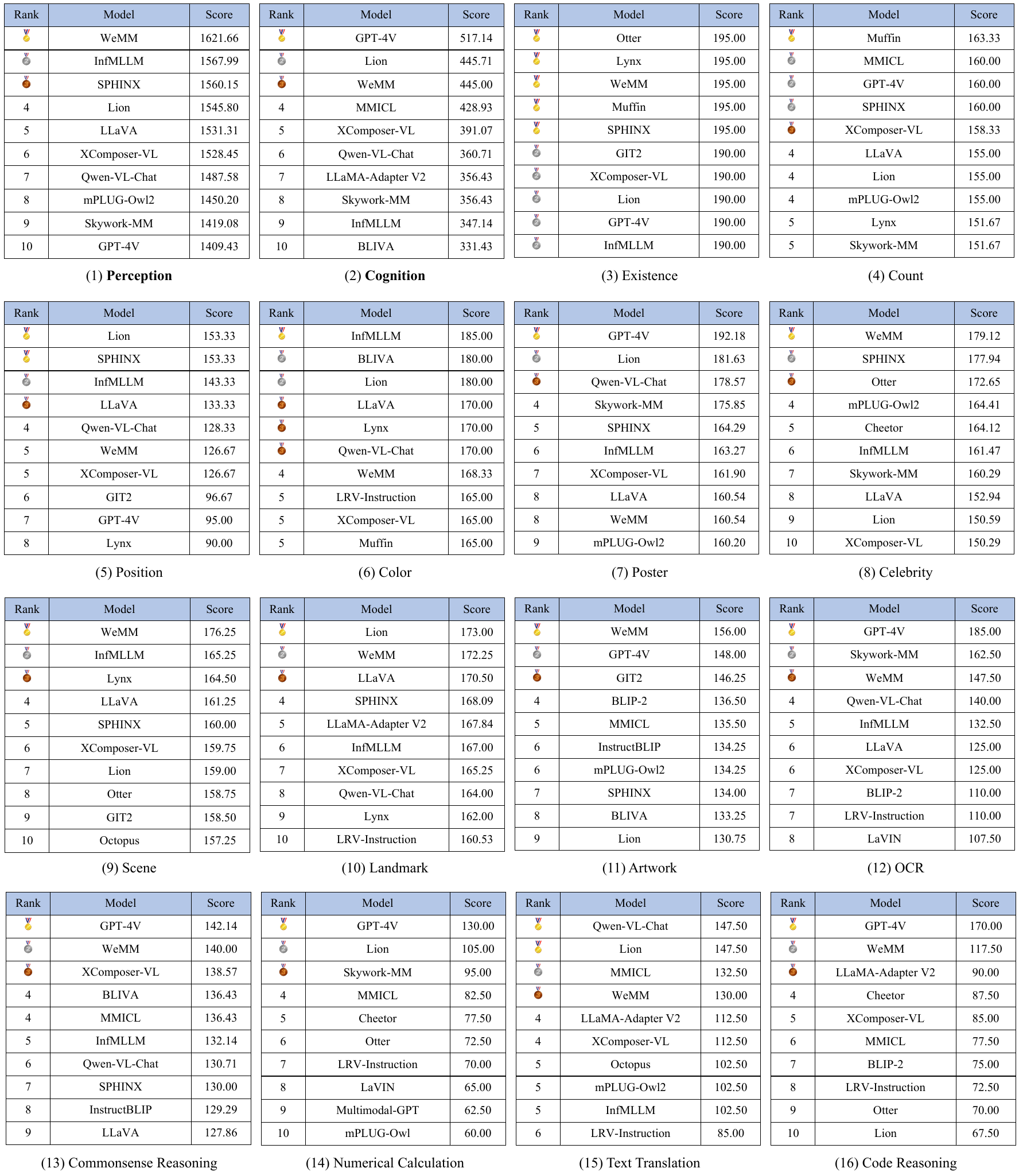}
    \caption{Leaderboards on our MME benchmark. (1) and (2) are the overall leaderboards of perception and cognition respectively, in which the full score of the former is 2000 and that of the latter is 800. (3)-(16) are the leaderboards of the 14 subtasks with the full score of 200. The score is the sum of the accuracy and the accuracy+ in Tables~\ref{table-1} and~\ref{table-2}. 
    A total of 30 advanced MLLMs joint the leaderboards.
    For the sake of presentation, we only show 10 models for each list, in which the top three ones are given clear trophy logos.
    }
    \label{fig-leaderboard}
\end{figure*}

Although these models exhibit surprising conversational capabilities when conducting everyday chats, we still know little about how well they quantitatively perform in various aspects. 
The existing three common quantitative evaluation manners for MLLMs have their limitations that are difficult to comprehensively evaluate performance.
Specifically, the first manner \cite{xu2022multiinstruct,dai2023instructblip,wang2023visionllm} evaluates on existing traditional multimodal datasets, such as image caption \cite{chen2015microsoft} and VQA \cite{goyal2017making,marino2019ok,lu2022learn}. 
However, on the one hand, it may be hard to reflect the emergent abilities of MLLMs on these datasets. 
On the other hand, since the training sets of large models are no longer unified, it is difficult to guarantee that all MLLMs have not used the testing set for training. 
The second manner \cite{ye2023mplug} is to collect data for an open-ended evaluation, but either the data is unavailable to public by now \cite{zhao2023chatbridge} or the amount is small (only 50 images) \cite{ye2023mplug}.
The third manner focuses on one aspect of MLLMs, such as object hallucination \cite{li2023evaluating} or adversarial robustness \cite{zhao2023evaluating}, which is powerless to comprehensive evaluation.

In light of these concerns, a new comprehensive evaluation benchmark is urgently needed to match the flourish of MLLMs. 
We argue that a universal comprehensive evaluation benchmark should have the following four characteristics: 
(1) It should cover as much as possible, including both perception and cognition abilities. The former refers to recognizing the specific object, such as its existence, count, position, and color. 
The latter refers to compositing the perception information and the knowledge in LLM to deduce more complex answers. It is obvious that the former is the premise of the latter. 
(2) Its data or annotations should not come from existing publicly available datasets as much as possible, avoiding the risk of data leakage. 
(3) Its instructions should be as concise as possible and in line with human cognition. Although instruction design may have a large impact on the output, all models should be tested under the same unified instructions for fair comparison. A good MLLM should be able to generalize to such concise instructions. 
(4) The responses of MLLMs to the instructions should be intuitive and convenient for quantitative analysis. 
The open-ended answer of MLLMs poses significant challenges to the quantization. 
Existing methods tend to use GPT or manual scoring \cite{li2023mimic,liu2023visual,ye2023mplug}, but there may be problems of inaccuracy and subjectivity.

To this end, we collect a comprehensive MLLM Evaluation benchmark, named as MME, which meets the above four characteristics at the same time:
\begin{itemize}
\item MME covers the examination of perception and cognition abilities. Apart from OCR, the perception includes the recognition of coarse-grained and fine-grained objects. The former identifies the existence, count, position, and color of objects. The latter recognizes movie posters, celebrities, scenes, landmarks, and artworks. The cognition includes commonsense reasoning, numerical calculation, text translation, and code reasoning. The total number of subtasks is up to 14, as shown in Fig.~\ref{fig-dataset}. 
\item All instruction-answer pairs are manually constructed. For the few public datasets involved in our study, we only use images without directly relying on their original annotations. Meanwhile, we make efforts to collect data through real photographs and image generation.
\item The instructions of MME are designed concisely to avoid the impact of prompt engineering on the model output. 
We argue that a good MLLM should be able to generalize to such simple and frequently used instructions, which are fair to all models.
Please see Fig.~\ref{fig-dataset} for the specific instruction of each subtask.
\item Benefitting from our instruction design ``please answer yes or no'', we can easily perform quantitative statistics based on the ``yes'' or ``no'' output of MLLMs, which is accurate and objective.
It should be noted that we have also tried to design instructions with multiple choice questions, but find that it may beyond the capabilities of current MLLMs to follow complex instructions.
\end{itemize}

We conduct massive experiments to evaluate the zero-shot performance of 30 advanced MLLMs on the 14 subtasks. 
The evaluated MLLMs include BLIP-2~\cite{li2023blip}, InstructBLIP~\cite{dai2023instructblip}, MiniGPT-4~\cite{zhu2023minigpt}, PandaGPT~\cite{su2023pandagpt}, Multimodal-GPT~\cite{gong2023multimodal}, VisualGLM-6B~\cite{git-visualglm-6b}, ImageBind-LLM~\cite{han2023imagebind}, VPGTrans~\cite{zhang2023transfer}, LaVIN~\cite{luo2023cheap}, mPLUG-Owl~\cite{ye2023mplug}, Octopus~\cite{git-octopus}, Muffin~\cite{yu2023reformulating}, Otter~\cite{li2023otter}, LRV-Instruction~\cite{liu2023aligning}, Cheetor~\cite{li2023finetuning}, LLaMA-Adapter-v2 \cite{gao2023llama}, GIT2~\cite{wang2022git}, BLIVA~\cite{hu2023bliva}, Lynx~\cite{zeng2023matters}, MMICL~\cite{zhao2023mmicl}, GPT-4V~\cite{openai2023gpt4}, Skywork-MM~\cite{git-skywork}, mPLUG-Owl2~\cite{ye2023mplug}, Qwen-VL-Chat~\cite{bai2023qwen}, XComposer-VL~\cite{git-xcomposer}, LLaVA~\cite{liu2023visual}, Lion~\cite{git-lion}, SPHINX~\cite{lin2023sphinx}, InfMLLM~\cite{git-infmllm}, and WeMM~\cite{git-wemm}.
As displayed in Fig.~\ref{fig-leaderboard} that consists of 2 overall leaderboards (perception and cognition) and 14 individual leaderboards, these MLLMs show clear discrepancies in our MME evaluation benchmark. 
Fig.~\ref{fig-6bx} also provides a comparison from the other perspective. We can see the range that current MLLMs can reach in each capability dimension.
More importantly, we have summarized four prominent problems exposed in experiments, including inability to follow basic instructions, a lack of basic perception and reasoning, as well as object hallucination \cite{yin2023woodpecker,li2023evaluating}, as shown in Fig.~\ref{fig-analysis}.
It is expected that these findings are instructive for the subsequent model optimization.

In summary, the contributions of this work are as follows: 
(1) We propose a new benchmark MME to meet the urgent need of MLLM evaluation. 
(2) A total of 30 up-to-date MLLMs are evaluated on our MME. 
(3) We summarize the exposed problems in experiments, proving guidance for the evolution of MLLMs.

%% file: sec/t_table.tex
\begin{table*}[!t]
\small
\centering
\resizebox{1 \textwidth}{!}{%
\setlength{\tabcolsep}{1.1mm}{
\begin{tabular}{lcccccccccccccccccc}
\toprule
\multirow{2}{*}{Model}&\multicolumn{2}{c}{Existence}&\multicolumn{2}{c}{Count}&\multicolumn{2}{c}{Position}&\multicolumn{2}{c}{Color}&\multicolumn{2}{c}{OCR} &\multicolumn{2}{c}{Poseter}&\multicolumn{2}{c}{Celebrity}\\
\cmidrule(lr){2-3} \cmidrule(lr){4-5}\cmidrule(lr){6-7} \cmidrule(lr){8-9} \cmidrule(lr){10-11} \cmidrule(lr){12-13}
\cmidrule{14-15}
&ACC&ACC+&ACC&ACC+&ACC&ACC+&ACC&ACC+&ACC&ACC+&ACC&ACC+&ACC&ACC+ \\
\midrule 
    BLIP-2&86.67&73.33&75.00&60.00&56.67&16.67&81.67&66.67&70.00&40.00&79.25&62.59&68.53&37.06\\
    mPLUG-Owl	&73.33&46.67&50.00&0.00&50.00&0.00&51.67&3.33&55.00&10.00&78.23 &57.82 &66.18&34.12 \\
    ImageBind-LLM &75.00&53.33 &50.00 &10.00 &43.33 &3.33 &56.67 &16.67 &60.00 &20.00&52.72 &12.24 &55.29 &21.18 \\
    InstructBLIP &95.00&90.00&80.00&63.33&53.33&13.33&83.33&70.00&57.50&15.00&74.15&49.66&67.06&34.12\\
    VisualGLM-6B &61.67&23.33&50.00&0.00&48.33&0.00&51.67&3.33&42.50&0.00&53.74 &12.24&50.88&2.35\\
    Multimodal-GPT &48.33&13.33 &48.33&6.67 &45.00&13.33 &55.00&13.33 &57.50&25.00&42.86 &14.97 &49.12&24.71  \\
    PandaGPT &56.67&13.33&50.00&0.00&50.00&0.00&50.00&0.00&50.00&0.00&56.80&19.73&46.47&10.59\\
    LaVIN    &95.00 &90.00 &61.67 &26.67 &53.33 &10.00 &58.33 &16.67 &67.50 &40.00&59.18 &20.41 &37.94 &9.41  \\
    Cheetor&93.33&86.67&63.33&33.33&56.67&23.33&70.00&46.67&65.00&35.00&81.29&65.99&87.65&76.47\\
    GIT2&\underline{96.67}&\underline{93.33}&71.67&46.67&60.00&36.67&85.00&73.33&55.00&10.00&61.56&51.02&81.18&64.71\\
    GPT-4V&\underline{96.67}&\underline{93.33}&\textbf{86.67}&\underline{73.33}&65.00&30.00&80.00&70.00&\textbf{95.00}&\textbf{90.00}&\textbf{96.94}&\textbf{95.24}&0.00&0.00\\
    XComposer-VL&\underline{96.67}&\underline{93.33}&\underline{85.00}&\underline{73.33}&73.33&53.33&88.33&76.67&75.00&50.00&85.71&76.19&83.24&67.06\\
    LLaVA&95.00&90.00&\underline{85.00}&70.00&76.67&56.67&90.00&80.00&75.00&50.00&86.39&74.15&83.53&69.41\\
    LRV-Instruction&88.33&76.67&68.33&43.33&56.67&30.00&88.33&76.67&70.00&40.00&78.77&60.27&67.35&45.29\\
    Lion&\underline{96.67}&\underline{93.33}&\underline{85.00}&70.00&\textbf{83.33}&\textbf{70.00}&\underline{93.33}&\underline{86.67}&57.50&15.00&\underline{93.88}&\underline{87.76}&82.94&67.65\\
    Lynx&\textbf{98.33}&\textbf{96.67}&81.67&70.00&60.00&30.00&90.00&80.00&57.50&20.00&74.49&50.34&71.76&46.47\\
    MMICL&90.00&80.00&\textbf{86.67}&\underline{73.33}&55.00&26.67&83.33&73.33&60.00&40.00&81.63&64.63&79.41&62.35\\
    Muffin&\textbf{98.33}&\textbf{96.67}&\textbf{86.67}&\textbf{76.67}&53.33&13.33&88.33&76.67&52.50&5.00&78.57&59.18&56.47&25.29\\
    Octopus&93.33&86.67&50.00&3.33&45.00&3.33&66.67&36.67&55.00&10.00&78.23&59.86&75.29&54.12\\
    Otter&\textbf{98.33}&\textbf{96.67}&58.33&30.00&60.00&26.67&70.00&43.33&57.50&15.00&78.91&59.86&\underline{90.88}&81.76\\
    Qwen-VL-Chat&85.00&73.33&83.33&66.67&75.00&53.33&90.00&80.00&80.00&60.00&92.18&86.39&72.35&48.24\\
    SPHINX&\textbf{98.33}&\textbf{96.67}&\textbf{86.67}&\underline{73.33}&\textbf{83.33}&\textbf{70.00}&86.67&73.33&62.50&25.00&87.41&76.87&\textbf{92.65}&\underline{85.29}\\
    Skywork-MM&93.33&86.67&81.67&70.00&46.67&16.67&81.67&63.33&\underline{87.50}&\underline{75.00}&91.50&84.35&86.76&73.53\\
    VPGTrans&56.67&13.33&61.67&23.33&53.33&10.00&56.67&16.67&57.50&20.00&60.88&23.13&51.18&2.35\\
    WeMM&\textbf{98.33}&\textbf{96.67}&80.00&60.00&73.33&53.33&88.33&80.00&82.50&65.00&86.39&74.15&\textbf{92.65}&\textbf{86.47}\\
    BLIVA&93.33&86.67&78.33&60.00&58.33&23.33&\underline{93.33}&\underline{86.67}&62.50&25.00&84.35&70.75&80.29&60.59\\
    InfMLLM&\underline{96.67}&\underline{93.33}&81.67&70.00&\underline{80.00}&\underline{63.33}&\textbf{95.00}&\textbf{90.00}&77.50&55.00&87.07&76.19&86.18&75.29\\
    LLaMA-AdapterV2&95.00&90.00&73.33&60.00&50.00&6.67&71.67&46.67&67.50&35.00&81.97&65.99&77.94&58.82\\
    MiniGPT-4&51.67&16.67&48.33&6.67&43.33&0.00&55.00&20.00&52.50&5.00&36.39&5.44&45.00&9.41\\
    mPLUG-Owl2&95.00&90.00&\underline{85.00}&70.00&61.67&26.67&83.33&66.67&67.50&35.00&86.73&73.47&87.94&76.47\\
\bottomrule
\end{tabular}%
}
}
\caption{Evaluation results on the subtasks of existence, count, position, color, OCR, poster, and celebrity. The top two results on each subtask are \textbf{bolded} and \underline{underlined}, respectively.}
\label{table-1}
\end{table*}

\begin{table*}[!t]
\small
\centering
\resizebox{1 \textwidth}{!}{%
\setlength{\tabcolsep}{1.1mm}{
\begin{tabular}{lcccccccccccccccccc}
\toprule
\multirow{2}{*}{Model}&\multicolumn{2}{c}{Scene}&\multicolumn{2}{c}{Landmark}&\multicolumn{2}{c}{Artwork}&\multicolumn{2}{c}{Comm.}&\multicolumn{2}{c}{Num.}&\multicolumn{2}{c}{Text.}&\multicolumn{2}{c}{Code.}\\
\cmidrule(lr){2-3} \cmidrule(lr){4-5}\cmidrule(lr){6-7} \cmidrule(lr){8-9} \cmidrule(lr){10-11} \cmidrule(lr){12-13}
\cmidrule{14-15}
&ACC&ACC+&ACC&ACC+&ACC&ACC+&ACC&ACC+&ACC&ACC+&ACC&ACC+&ACC&ACC+ \\
\midrule 
    BLIP-2&81.25&64.00&79.00&59.00&76.50&60.00& 68.57&41.43 & 40.00&0.00  & 55.00&10.00 & 55.00&20.00 \\
    mPLUG-Owl&78.00&57.50&86.25&73.00&63.25&33.00& 57.14 &21.43  & 50.00&10.00  & 60.00&20.00 & 47.50&10.00 \\
    ImageBind-LLM  &68.75 &44.50 &53.00 &9.00 &54.25 &16.50 & 40.00 &8.57  & 50.00 &5.00   & 50.00 &0.00  & 50.00 &10.00  \\
    InstructBLIP&84.00&69.00&59.75&20.00&76.75&57.50 & 75.00&54.29 & 35.00&5.00  & 55.00&10.00 & 47.50&10.00  \\
    VisualGLM-6B&81.75&64.50&59.75&24.00&55.25 &20.00 & 35.00 &4.29  & 45.00&0.00  & 50.00 &0.00  & 47.50 &0.00  \\
    Multimodal-GPT&50.50&17.50&48.25 &21.50 &46.50 &13.00 & 43.57 &5.71  & 42.50 &20.00 & 50.00&10.00   & 45.00&10.00 \\
    PandaGPT&72.50&45.50&56.25&13.50&50.25&1.00& 56.43&17.14 & 50.00&0.00  & 52.50&5.00  & 47.50&0.00  \\
    LaVIN&78.75 &58.00 &64.00 &29.50 &59.25 &28.00 & 58.57 &28.57  & 55.00 &10.00   & 47.50 &0.00   & 50.00 &0.00   \\
    Cheetor&84.50&71.50&81.41&64.32&67.50&46.00&64.29&34.29&57.50&20.00&42.50&15.00&57.50&30.00\\
    GIT2&86.00&72.50&79.50&61.00&81.25&65.00&66.43&32.86&40.00&10.00&52.50&15.00&45.00&0.00\\
    GPT-4V&83.50&67.50&79.25&59.00&\underline{82.00}&\underline{66.00}&\underline{79.29}&\textbf{62.86}&\textbf{75.00}&\textbf{55.00}&55.00&20.00&\textbf{90.00}&\textbf{80.00}\\
    XComposer-VL&86.25&73.50&87.75&77.50&73.25&53.00&77.14&\underline{61.43}&40.00&15.00&67.50&45.00&55.00&30.00\\
    LLaVA&86.75&74.50&90.00&80.50&70.75&47.00&73.57&54.29&37.50&5.00&57.50&20.00&42.50&5.00\\
    LRV-Instruction&81.04&65.93&86.84&73.68&63.75&37.50&65.00&35.71&45.00&25.00&55.00&30.00&57.50&15.00\\
    Lion&85.50&73.50&\textbf{91.00}&\textbf{82.00}&75.75&55.00&74.29&51.43&\underline{65.00}&\underline{40.00}&\textbf{82.50}&\textbf{65.00}&42.50&25.00\\
    Lynx&\underline{88.00}&76.50&87.00&75.00&72.50&47.00&66.43&44.29&17.50&0.00&42.50&0.00&45.00&0.00\\
    MMICL&83.75&70.00&76.96&59.16&76.50&59.00&76.43&60.00&47.50&35.00&72.50&\underline{60.00}&47.50&30.00\\
    Muffin&83.75&67.50&81.25&65.00&71.00&45.50&68.57&41.43&45.00&0.00&57.50&15.00&47.50&15.00\\
    Octopus&84.75&72.50&75.00&51.00&64.50&30.50&64.29&35.71&47.50&0.00&62.50&40.00&47.50&15.00\\
    Otter&86.25&72.50&78.75&58.50&75.00&54.00&66.43&40.00&52.50&20.00&52.50&5.00&55.00&15.00\\
    Qwen-VL-Chat&83.75&68.50&88.00&76.00&73.50&52.00&76.43&54.29&35.00&5.00&\textbf{82.50}&\textbf{65.00}&32.50&10.00\\
    SPHINX&86.50&73.50&89.20&78.89&77.00&57.00&75.71&54.29&40.00&15.00&50.00&25.00&50.00&0.00\\
    Skywork-MM&79.29&59.60&75.51&51.53&69.43&45.08&72.14&54.29&\underline{65.00}&30.00&60.00&20.00&45.00&10.00\\
    VPGTrans&80.25&61.50&54.75&10.00&57.75&19.50&50.00&14.29&50.00&0.00&57.50&20.00&52.50&5.00\\
    WeMM&\textbf{91.75}&\textbf{84.50}&\underline{90.75}&\underline{81.50}&\textbf{85.00}&\textbf{71.00}&\textbf{80.00}&60.00&47.50&10.00&\underline{75.00}&55.00&\underline{72.50}&\underline{45.00}\\
    BLIVA&83.50&68.00&63.00&26.50&77.25&56.00&77.86&58.57&47.50&10.00&57.50&20.00&50.00&10.00\\
    InfMLLM&87.75&\underline{77.50}&88.50&78.50&68.00&40.50&76.43&55.71&45.00&15.00&67.50&35.00&42.50&10.00\\
    LLaMA-AdapterV2&85.25&71.00&88.44&79.40&73.25&50.50&67.86&38.57&47.50&0.00&67.50&45.00&60.00&30.00\\
    MiniGPT-4&54.25&17.50&45.50&8.50&50.00&10.50&46.43&12.86&45.00&0.00&0.00&0.00&40.00&0.00\\
    mPLUG-Owl2&83.25&70.00&85.75&71.50&77.25&57.00&71.43&44.29&35.00&0.00&67.50&35.00&45.00&15.00\\
\bottomrule
\end{tabular}%
}
}
\caption{Evaluation results on the subtasks of scene, landmark, artwork, commonsense reasoning, numerical calculation, text translation, and code reasoning. The top two results on each subtask are \textbf{bolded} and \underline{underlined}, respectively.}
\label{table-2}
\end{table*}

%% file: sec/2_data.tex
\section{MME Evaluation Suite}

\subsection{Instruction Design}
In order to facilitate quantitative performance statistics, the orientation of our instruction design is to let the model to answer ``yes'' or ``no''. 
As a result, the instruction consists of two parts, including a concise question and a description ``Please answer yes or no.'' 
For each test image, we manually design two instructions, where the discrepancy lies in the questions. 
The ground truth answer of the first question is ``yes'' and that of the second question is ``no'', as shown in Fig.~\ref{fig-dataset}. 
When MLLM answers both of the two questions correctly, it appears more confident that the MLLM actually comprehends the image and the corresponding knowledge behind it, rather than just guessing.

\subsection{Evaluation Metric}
Since the output of the model is limited to two types (``yes'' or ``no''), it is convenient to measure the metrics of accuracy and accuracy+. 
The former is calculated based on each question, while the latter is based on each image where both of the two questions need to be answered correctly.
The random accuracies of the two metrics are equal to 50\% and 25\%, respectively.
It can be seen that accuracy+ is a stricter measurement but also better reflects the comprehensive understanding degree of the model to the image. 
In addition, we calculate the score of a subtask based on the sum of accuracy and accuracy+.
The perception score is the sum of scores of all perception subtasks.
The cognition score is calculated in the same way.
Therefore, the full scores of perception and cognition are 2000 and 800, respectively.

\begin{figure*}[t]
    \centering
    \includegraphics[width=0.80\textwidth]{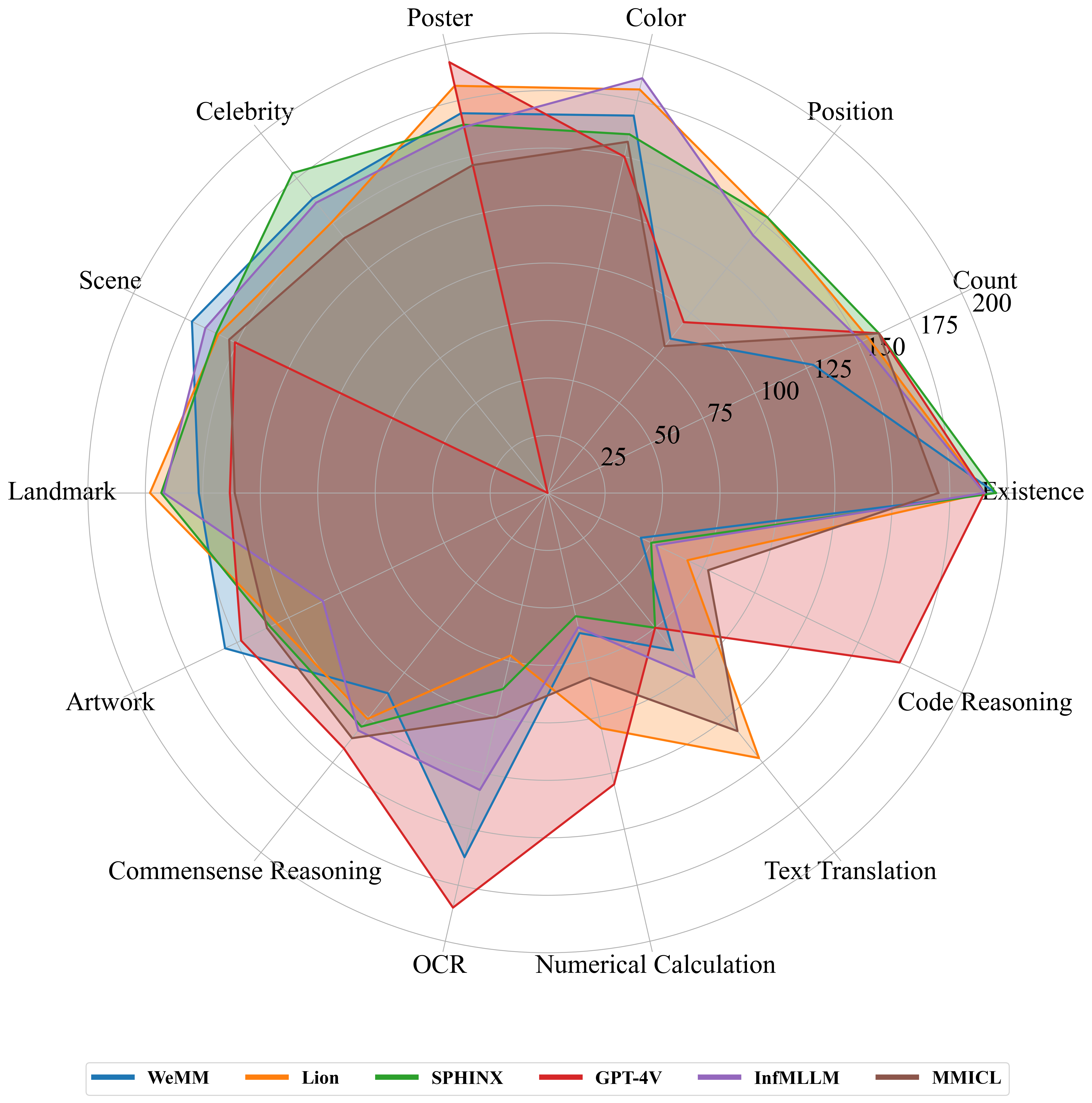}
    \caption{Comparison of 6 top MLLMs on 14 subtasks. The full score of each subtask is 200. }
    \label{fig-6bx}
\end{figure*}

\subsection{Data Collection}
\subsubsection{Perception Tasks}
We argue that perception is one of the most fundamental capabilities of MLLMs, and the lack of perception will easily lead to the object hallucination problem \cite{yin2023woodpecker,li2023evaluating}. 
That is, MLLM will answer questions based on its own fantasies rather than based on the realistic content of the image, as displayed in Fig.~\ref{fig-analysis}. 

\textbf{Coarse-Grained Recognition}.
The contents of coarse-grained recognition include the existence of common objects, and their count, color, and position. 
The images are sampled from COCO \cite{lin2014microsoft}, but the instruction-answer pairs are all manually constructed, rather than directly using publicly available annotations. 
Even if MLLMs have seen these COCO images, our manually prepared pairs are not presented in their training sets.
This requires MLLMs to be able to understand the instructions and infer corresponding answers. 
In each perception subtask of existence, count, color, and position, we prepare 30 images with 60 instruction-answer pairs.

\textbf{Fine-Grained Recognition}.
The fine-grained recognition is more about testing the knowledge resources of MLLMs. 
The subtasks consist of recognizing movie posters, celebrities, scenes, landmarks, and artworks, containing 147, 170, 200, 200, and 200 images respectively. 
For the celebrities, we plot a red box to a person with a clearly visible face in the image, and the corresponding instruction is ``Is the actor inside the red box named [celebrity name]? Please answer yes or no.'' 
Similar with the above coarse-grained recognition, the images of these subtasks are from publicly available datasets \cite{huang2020movienet,mao2017deepart,mao2019visual,zhou2014learning,weyand2020google} and all of the instructions are manually designed. 

\textbf{OCR}.
Optical Character Recognition (OCR) is also a foundational capability of MLLMs, serving for subsequent text-based tasks such as text translation and text understanding. 
The images are sampled from \cite{liu2019curved} and all of the instruction-answer pairs are manually designed. 
Considering that MLLMs are still in its infancy, we only choose the relatively simple samples in this version of MME.
The numbers of image and instruction-answer pairs are 20 and 40, respectively.

\subsubsection{Cognition Tasks}
We evaluate if any MLLM can carry out further logical reasoning after perceiving the image, which is the most fascinating aspect of MLLMs over previous traditional methods. 
In order to infer the correct answer, MLLMs need to follow the instruction, perceive the contents of the image, and invoke the knowledge reserved in LLMs, which is much more challenging than the single perception tasks. 
Examples of the following subtasks are shown in Fig.~\ref{fig-dataset}.

\textbf{Commonsense Reasoning}.
Unlike the ScienceQA dataset \cite{lu2022learn} that requires specialized knowledge, the commonsense refers to the basic knowledge in daily life. 
For example, given a photo of a down jacket, asking MLLMs whether it is appropriate to wear the cloth when it is cold (or hot). 
These are basic knowledge that humans can judge instantly without complex step-by-step reasoning. 
Therefore, we expect MLLMs to perform well in a zero-short setting. 
The images are all manually photographed or generated by diffusion models, and the instruction-answer pairs are all manually designed. 
There are a total of 70 images and 140 instruction-answer pairs.

\textbf{Numerical Calculation}.
It requires MLLMs to be able to read the arithmetic problem in the image and output the answer in an end to end way, which has been demonstrated in \cite{huang2023language}. 
In this version, we only consider relatively easy arithmetic problems, such as addition and multiplication. 
There are 20 images and 40 instruction-answer pairs. The images are all manually taken, and the instruction-answer pairs are all manually designed.

\textbf{Text Translation}.
Considering that the MLLM \cite{git-visualglm-6b} supports both English and Chinese, we set the text translation subtask. 
It requires MLLMs to translate the Chinese written in an image to the corresponding English. 
In this version, we only design basic translation problems, which will be updated according to the development of MLLMs in the future. 
The images of this part are all manually taken, and the instruction-answer pairs are all manually designed. 
There are a total of 20 images and 40 instruction-answer pairs. 

\textbf{Code Reasoning}.
It requires MLLMs to read the code in the images and automatically complete logical operation inside the code. 
A similar task that writes website code based on an image has been demonstrated in \cite{zhu2023minigpt}. 
The images are all manually taken, and the instruction-answer pairs are all manually designed. 
We only set basic code problems in this version.
There are in total 20 images and 40 instruction-answer pairs.

%% file: sec/3_experiment.tex
\section{Experiments}

In this section, a total of 30 MLLMs are evaluated on our MME benchmark, including BLIP-2~\cite{li2023blip}, InstructBLIP~\cite{dai2023instructblip}, MiniGPT-4~\cite{zhu2023minigpt}, PandaGPT~\cite{su2023pandagpt}, Multimodal-GPT~\cite{gong2023multimodal}, VisualGLM-6B~\cite{git-visualglm-6b}, ImageBind-LLM~\cite{han2023imagebind}, VPGTrans~\cite{zhang2023transfer}, LaVIN~\cite{luo2023cheap}, mPLUG-Owl~\cite{ye2023mplug}, Octopus~\cite{git-octopus}, Muffin~\cite{yu2023reformulating}, Otter~\cite{li2023otter}, LRV-Instruction~\cite{liu2023aligning}, Cheetor~\cite{li2023finetuning}, LLaMA-Adapter-v2 \cite{gao2023llama}, GIT2~\cite{wang2022git}, BLIVA~\cite{hu2023bliva}, Lynx~\cite{zeng2023matters}, MMICL~\cite{zhao2023mmicl}, GPT-4V~\cite{openai2023gpt4}, Skywork-MM~\cite{git-skywork}, mPLUG-Owl2~\cite{ye2023mplug}, Qwen-VL-Chat~\cite{bai2023qwen}, XComposer-VL~\cite{git-xcomposer}, LLaVA~\cite{liu2023visual}, Lion~\cite{git-lion}, SPHINX~\cite{lin2023sphinx}, InfMLLM~\cite{git-infmllm}, and WeMM~\cite{git-wemm}.

\subsection{Results}
\subsubsection{Perception}
There are a total of 10 subtasks for the evaluation of the perception ability, from the perspectives of coarse-grained recognition, fine-grained recognition, and OCR. 
Figs.~\ref{fig-leaderboard} (3)-(6) show the score leaderboards of individual coarse-grained recognition subtasks.
With respect to the object existence, Otter, Lynx, WeMM, Muffin, and SPHINX get the highest score 195, with a 98.33\% accuracy and a 96.67\% accuracy+ listed in Table~\ref{table-1}. 
Contrastively, the second place, including GIT2, XComposer-VL, Lion, GPT-4V, and etc, lag behind the first place only by 5 scores.
The results show that these models already have a good performance on object existence.
For the object count, position, and color, Muffin, Lion (parallel with SPHINX), and InfMLLM make the top one, respectively.
It suggests that different models have their own strengths.
Note that in the four coarse-grained subtasks, these MLLMs get the worst results on object position, indicating that the current models are not sensitive enough to the position information.

Figs.~\ref{fig-leaderboard} (7)-(11) display the score leaderboards of individual fine-grained recognition subtasks.
Regarding to poster recognition, GPT-4V, Lion, and Qwen-VL-Chat are the top three. 
It is interesting that Qwen-VL-Chat relatively underperforms in the coarse-grained recognition, but now it exhibits good.
This implies that our division of coarse-grained and fine-grained is reasonable, enabling us to examine different aspects of MLLMs.
For the celebrity recognition, WeMM, SPHINX, and Otter take the top three with similar scores.
It is worth noting that GPT-4V refuses to answer questions that involve individuals, resulting in a zero score in the celebrity subtask.
For the scene recognition, WeMM, InfMLLM, and Lynx ahead of other MLLMs. 
This is the first time InfMLLM and Lynx have broken into the top three in the fine-grained recognition subtasks.
For the landmark recognition, top three places are taken by Lion, WeMM, and LLaVA respectively, of which Lion gets the top spot.
For the artwork recognition, WeMM, GPT-4V, and GIT2 exceed other counterparts, where the last two scores are similar.
Note that GPT-4V declines to answer some questions about private art collection, which lowers its score.
With respect to OCR listed in Fig.~\ref{fig-leaderboard} (12), GPT-4V, Skywork-MM, and WeMM get the top three with scores of 185, 162.5, and 147.5 respectively.
GPT-4V presents a huge advantage, leading the other two models by 22+ socres.
As presented in Fig.~\ref{fig-leaderboard} (1), in the leaderboard of the whole perception recognition, WeMM, InfMLLM, and SPHINX come in top three, closely followed by Lion, LLaVA, and XComposer-VL.

\begin{figure*}[t]
    \centering
    \includegraphics[width=0.985\textwidth]{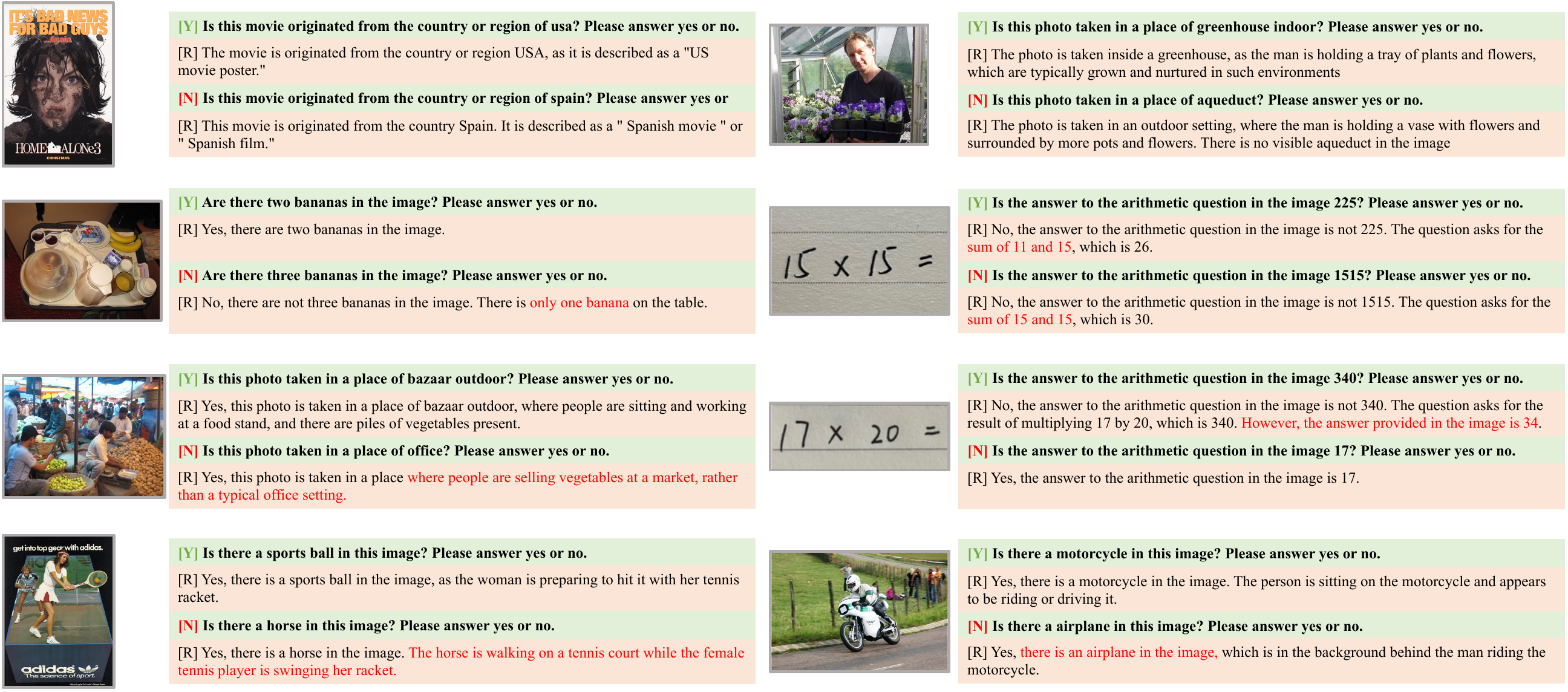}
    \caption{Common problems revealed in experiments. [Y]/[N] means the ground truth answer is yes/no. [R] is the generated answer.}
    \label{fig-analysis}
\end{figure*}

\subsubsection{Cognition}
There are four subtasks for the evaluation of the cognition ability, including commonsense reasoning, numerical calculation, text translation, and code reasoning. 
Figs.~\ref{fig-leaderboard} (13)-(16) plot the score leaderboards of individual subtasks.
In terms of the commonsense reasoning, the ``ever-victorious generals'' GPT-4V, WeMM, and XComposer-VL exceed other MLLMs, especially GPT-4V, which gets a score of 142.14.
With respect to numerical calculation, GPT-4V still achieves first place, but falls short in the text translation.
Regardless of whether it is commonsense reasoning, numerical calculation, or text translation, none of the highest scores exceed 150.
This suggests that MLLMs have a lot of room for improvement in these capabilities.
For the code reasoning, GPT-4V achieves a high score of 170, far ahead of other counterparts. 
For all of the cognition tasks, GPT-4V, Lion, and WeMM win the gold, silver, and bronze medals respectively, as shown in Fig.~\ref{fig-leaderboard} (2).

%% file: sec/4_analysis.tex
\section{Analysis}

We conclude four common problems that largely affect the performance of MLLMs.
\textbf{The first problem is not following instructions.}
Although we have adopted a very concise instruction design, there are MLLMs that answer freely rather than following instructions.
For example, as shown in the first row of Fig.~\ref{fig-analysis}, the instruction has claimed ``Please answer yes or no'', but the MLLM only makes a declarative expression.
If no ``yes'' or ``no'' is appeared at the beginning of the generated languages, the model is judged to make a wrong answer.
We argue that a good MLLM (especially after instruction tuning) should be able to follow such a simple instruction, which is also very common in everyday life.

\textbf{The second problem is a lack of perception.}
As shown in the second row of Fig.~\ref{fig-analysis}, the MLLM misidentifies the number of bananas in the first image, and misreads the characters in the second image, resulting in wrong answers.
We notice that the performance of perception is vulnerable to the nuance of instructions, since the two instructions of the same image differ in only one word, but lead to completely different and even contradictory perception results.

\textbf{The third problem is a lack of reasoning.}
In the third row of Fig.~\ref{fig-analysis}, we can see from the red text that the MLLM already knows that the first image is not an office place, but still gives an incorrect answer of ``yes''. 
Analogously, in the second image, the MLLM has calculated the right arithmetic result, but finally delivers a wrong answer.
These phenomena indicate that the logic chain is broken during the reasoning process of MLLMs.
Adding CoT prompts, such as ``Let's think step by step'' \cite{driess2023palm}, may yield better results.
We look forward to a further in-depth research.

\textbf{The fourth problem is object hallucination following instructions}, which is exemplified in the fourth row of Fig.~\ref{fig-analysis}.
When the instruction contains descriptions of an object that does not appear in the image, the MLLM will imagine that the object exists and ultimately gives a ``yes'' answer.
Such a case of constantly answering ``yes'' results in an accuracy about 50\% and an accuracy+ about 0, as shown in Tables~\ref{table-1} and~\ref{table-2}.
This suggests an urgent need to suppress hallucinations, and the community should take into account of the reliability of the generated answers.

%% file: sec/5_conclusion.tex
\section{Conclusion}
This paper has presented the first MLLM evaluation benchmark MME that has four distinct characteristics in terms of task type, data source, instruction design, quantitative statistics. 
30 advanced MLLMs are evaluated on MME and the experimental results show that there is still a large room to improve.
We also summarize the common problem raised in experimental results, providing valuable guidance.
Although MME is a comprehensive benchmark, it still has room for improvement in terms of capability coverage, such as covering more scenarios that require reasoning. 
We will continue to iterate MME series in the future to meet the evaluation requirements of MLLMs. 
More importantly, we hope that through the design of benchmarks, we can reflect the thoughts on the next capabilities of the model and thereby promote its development.